\pdfoutput=1

\documentclass[11pt]{article}

\usepackage[final]{acl}

\usepackage{times}
\usepackage{latexsym}

\usepackage[T1]{fontenc}

\usepackage[utf8]{inputenc}

\usepackage{microtype}

\usepackage{inconsolata}

\usepackage{graphicx}
\usepackage{bm}
\usepackage{amsfonts}
\usepackage{amsmath}
\usepackage{booktabs}
\usepackage{multirow}
\usepackage{subfigure}
\usepackage{CJKutf8}
\usepackage{utfsym}
\usepackage{color}
\title{A Non-autoregressive Generation Framework for End-to-End Simultaneous Speech-to-Speech Translation }

\author{Zhengrui Ma$^{1,3}$, Qingkai Fang$^{1,3}$, Shaolei Zhang$^{1,3}$, Shoutao Guo$^{1,3}$\\
 {\bf Yang Feng$^{1,2,3}$\thanks{$\;\;$Corresponding author: Yang Feng},} {\bf Min Zhang$^{4}$} \\
\textsuperscript{\rm1}Key Laboratory of Intelligent Information Processing \\ Institute of Computing Technology, Chinese Academy of Sciences \\
\textsuperscript{\rm2}Key Laboratory of AI Safety, Chinese Academy of Sciences \\
\textsuperscript{\rm3}University of Chinese Academy of Sciences\\
\textsuperscript{\rm4}School of Future Science and Engineering, Soochow University \\
{$\;\:$\texttt{\{\href{mailto:mazhengrui21b@ict.ac.cn}{mazhengrui21b},\href{mailto:fengyang@ict.ac.cn}{fengyang}\}@ict.ac.cn}}
{$\;\:$\texttt{\href{mailto:zhangminmt@hotmail.com}{zhangminmt}@hotmail.com}}}

\begin{document}
\maketitle
\begin{abstract}

Simultaneous translation models play a crucial role in facilitating communication. However, existing research primarily focuses on text-to-text or speech-to-text models, necessitating additional cascade components to achieve speech-to-speech translation. These pipeline methods suffer from error propagation and accumulate delays in each cascade component, resulting in reduced synchronization between the speaker and listener. To overcome these challenges, we propose a novel non-autoregressive generation framework for simultaneous speech translation (NAST-S2$x$\footnote{\enspace $x \in \{\mathrm{text}, \mathrm{speech} \}$}), which integrates speech-to-text and speech-to-speech tasks into a unified end-to-end framework.
We develop a non-autoregressive decoder capable of concurrently generating multiple text or acoustic unit tokens upon receiving fixed-length speech chunks. The decoder can generate blank or repeated tokens and employ CTC decoding to dynamically adjust its latency. 
Experimental results show that NAST-S2$x$ outperforms state-of-the-art models in both speech-to-text and speech-to-speech tasks. It achieves high-quality simultaneous interpretation within a delay of less than 3 seconds and provides a 28× decoding speedup in offline generation.\footnote{\textcolor{magenta}{\enspace \textbf{Project}: \texttt{\hypersetup{urlcolor=magenta}\href{https://github.com/ictnlp/NAST-S2x}{https://github.com/ictnlp/NAST-S2x}}}}

\end{abstract}

\section{Introduction}
Simultaneous machine translation \citep{DBLP:journals/corr/ChoE16,gu-etal-2017-learning, pmlr-v70-raffel17a,ma-etal-2019-stacl,arivazhagan-etal-2019-monotonic} models are widely applied in communication scenarios, eliminating barriers between individuals with different linguistic backgrounds. In practice, simultaneous translation systems can be broadly categorized into speech-to-text \citep[Simul-S2T;][]{ma-etal-2020-simulmt} and speech-to-speech \citep[Simul-S2S;][]{zheng-etal-2020-fluent} variants. 
Regardless of the modality of output, simultaneous translation models initiate generation before receiving the complete input to maintain synchrony between the listener and speaker. This necessitates models to delicately balance between translation quality and latency.

\begin{figure}[t]
\centering
\includegraphics[width=2.8in]{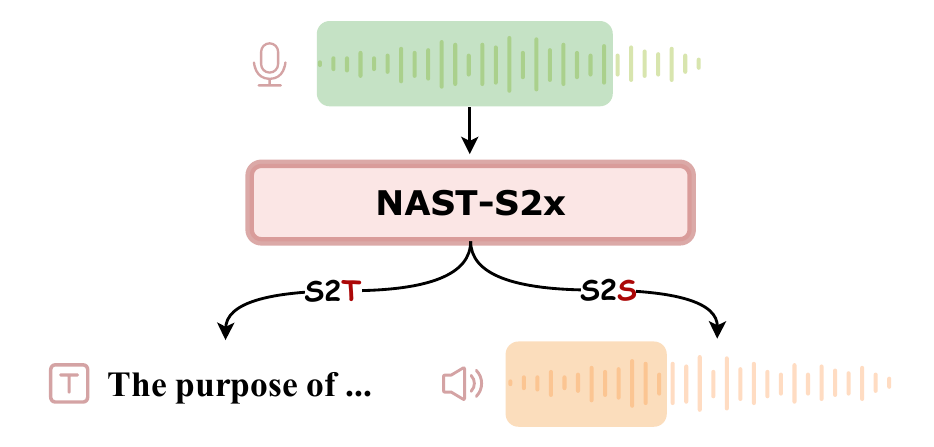}
\caption{NAST-S2$x$ can perform both Simul-S2T and Simul-S2S tasks within a unified end-to-end framework. The model generates speech output directly without the need to produce intermediate target text tokens}
\label{overview}
\vspace{-5mm}
\end{figure}

Most research on simultaneous machine translation  primarily focuses on either text-to-text \citep{Ma2020Monotonic,miao-etal-2021-generative} or speech-to-text models \citep{tang-etal-2023-hybrid,NEURIPS2023_8df70595}, necessitating additional cascaded components such as streaming automatic speech recognition \citep{chiu*2018monotonic,9053896} and incremental text-to-speech synthesis \citep{ma-etal-2020-incremental} for achieving speech-to-speech interpretation \citep{zheng-etal-2020-fluent}. However, pipeline methods often suffer from error propagation and delay accumulation. The intermediate texts serve as information bottlenecks, hindering subsequent cascade components from accessing the original information and correcting errors. Moreover, each component operates with independent streaming strategies, resulting in cumulative delays thus diminishing synchronization between the speaker and listener. Given these challenges, the emergence of end-to-end Simul-S2S models has garnered increasing attention in the research community.

Recent success of end-to-end offline speech-to-speech translation (Offline-S2S) has paved the way for the development of end-to-end Simul-S2S models. Particularly, \citet{lee-etal-2022-direct} construct a direct speech-to-unit model (S2UT), which predicts self-supervised discrete representations of target speech. Waveforms are subsequently generated using a separate unit-based vocoder \citep{polyak21_interspeech}. 
On this basis, \citet{ma2022direct} builds the first end-to-end Simul-S2S model by introducing a variational version of monotonic multihead attention \citep{Ma2020Monotonic}.
However, previous works are mainly limited to predicting units in an autoregressive manner, which is suboptimal for end-to-end Simul-S2S models. Considering that the acoustic unit sequence is 25 times longer than the corresponding text sequence on average, autoregressive unit prediction often leads to issues such as hallucination or truncation \citep{seamless2023}. Moreover, the sequential prediction of long unit sequences imposes a significant computational time overhead, making it impractical for delay-sensitive Simul-S2S systems. To tackle these challenges, our focus is on developing a non-autoregressive end-to-end Simul-S2S model, aiming for enjoying the merits of an end-to-end system without the necessity of intermediate text decoding, while benefiting from the efficiency inherent in non-autoregressive generation.

In this work, we propose a non-autoregressive generation framework for end-to-end simultaneous speech-to-any translation (NAST-S2$x$). Inspired by recent advances in non-autoregressive generation \citep{NEURIPS2022_35f805e6,ma-etal-2023-non}, we develop a non-autoregressive decoder capable of concurrently generating multiple text or acoustic unit tokens upon receiving each fixed-length speech chunk. The entire generation adopts a chunk-to-chunk approach, while avoiding the unstable expected training method \citep{NEURIPS2023_8df70595}. The model can produce blank or repeated tokens and perform CTC decoding \citep{10.1145/1143844.1143891} to adjust its latency dynamically. Considering the difficulty of the speech translation task and aiming to leverage intermediate text data to assist training, we further introduce a two-step glancing and a multi-task non-monotonic training strategy, which largely enhances the translation performance while maintaining the end-to-end nature of our model.

Extensive experiments highlight the superiority of our NAST-S2$x$. In Simul-S2T, its performance is on par with state-of-the-art models. In Simul-S2S, it significantly surpasses cascade Simul-S2T + TTS baselines, achieving high-quality simultaneous interpretation within a delay of less than 3 seconds. In Offline-S2S, it matches the performance of the strong autoregressive baseline while providing a 28× inference speedup.

\section{Preliminaries}

\subsection{Simultaneous Speech Translation}
Simultaneous speech translation models often process a streaming sequence of acoustic features $\bm{x} = \{x_1,...,x_m \} $ as input, extracted from speech samples every $T_w$ ms.
Simultaneous translation models can be further categorized into speech-to-text (Simul-S2T) and speech-to-speech (Simul-S2S) variants based on the output modality.

\subsubsection{Simul-S2T}
A Simul-S2T model generates a translated text sequence $\bm{y} = \{y_1,...,y_n \}$ in a streaming fashion. To quantify the extent of source information taken into account during the generation, a monotonic non-decreasing function $g(t)$ is introduced to represent the number of observed frames when generating $y_t$. 

To assess the latency of Simul-S2T models, \citet{ma-etal-2020-simulmt} introduce a modified version of average lagging \citep[\texttt{AL};][]{ma-etal-2019-stacl} for speech-to-text task. They measure the lagging based on time instead of steps, and the metric is defined as:
\begin{equation}
    AL = \frac{1}{\tau(|\bm{x}|)} \sum^{\tau(|\bm{x}|)}_{t=1}{d(t)-\frac{|\bm{x}|}{|\bm{y}^*|}\cdot T_w \cdot (t-1)},
\end{equation}
where $|\bm{x}|$ and $|\bm{y}^*|$ represent the lengths of source frames and reference text. $\tau(|\bm{x}|)$ is the index of the first generated token when the source is complete, and $d(t)$ is the delay of generating $y_t$. \citet{ma-etal-2020-simulmt} further defines computation-aware and non-computation-aware versions of $d(t)$. The former, $d_{CA}(t)$, is defined as the elapsed time from the beginning of the whole process, while the latter is simply calculated as $d_{NCA}(t) = g(t) \cdot T_w$. As the non-computation-aware metric is independent of implementation, most previous studies adopt this metric for comparisons, focusing on the algorithm.

\subsubsection{Simul-S2S}
A Simul-S2S model further synthesizes translated text into speech. To assess the translation quality of Simul-S2S models, a separate offline automatic speech recognition system is employed to transcribe the generated speech $\mathcal{Y}(t)$ into text $\bm{y}$ for computing \texttt{ASR-BLEU} against the reference \citep{DBLP:conf/interspeech/JiaWBMJCW19}. To evaluate latency, a forced aligner is usually introduced to align the transcription $\bm{y}$ with $\mathcal{Y}(t)$ to acquire the delay of each token in $\bm{y}$. 
Subsequently, the \texttt{AL} metric, as defined in Simul-S2T, can be calculated for $\bm{y}$ \citep{ma2022direct}.\footnote{A detailed description of the latency metric used in our Simul-S2S experiments is provided in Section \ref{sec:exp:s2s}.}

\subsection{Speech-to-Unit Translation}
Recent success of self-supervised representation learning in speech has opened up a new avenue for building speech-to-speech translation systems. The discretized units derived from clustering speech representations allow models to predict speech in a manner analogous to text. \citet{lee-etal-2022-direct} build the first speech-to-unit (S2UT) translation model with autoregressive Transformer \citep{NIPS2017_3f5ee243}. They utilize a HuBERT \citep{9585401} pre-trained on an unlabelled speech corpus and perform $k$-means algorithm to the learned representations of each 20ms chunk to produce $K$ cluster centroids. Each chunk is then assigned the index of its nearest centroid serving as the label. Consequently, a target utterance can be encoded as a sequence of cluster indices $\bm{z}=\{z_1,z_2,...,z_T\}$, $z_i \in \{0,1,...,K-1\}$, $\forall 1 \leq i \leq T$, where $T$ is the number of chunks. S2UT model can be trained using cross-entropy. A separate unit-based vocoder \citep{polyak21_interspeech} is employed to convert the predicted acoustic unit sequence into waveform.

\section{Approach}
We provide a detailed introduction to our non-autoregressive generation framework for end-to-end simultaneous speech-to-any translation in this section.
\begin{figure*}[ht]
\centering
\includegraphics[width=6.0in]{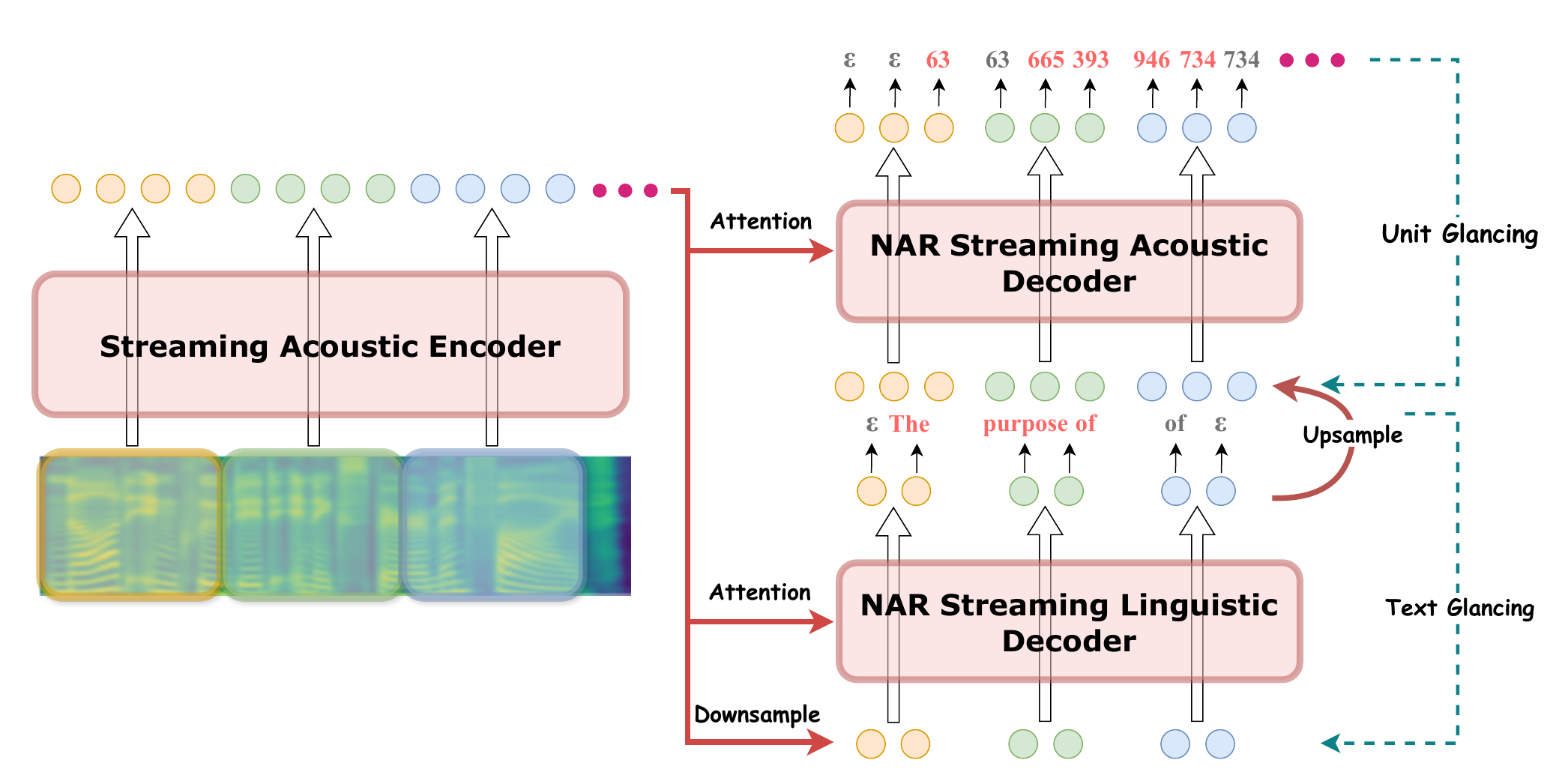}
\caption{Overview of the proposed non-autoregressive generation framework for end-to-end simultaneous speech-to-any translation (NAST-S2$x$, $x \in \{\mathrm{text}, \mathrm{speech} \}$). Different colors indicate different chunks.} 
\label{architecture}
\end{figure*}
\subsection{Architecture}
As illustrated in Figure \ref{architecture}, NAST-S2$x$ consists of a chunk-based acoustic streaming encoder and a chunk-based non-autoregressive (NAR) streaming decoder. This non-autoregressive decoder comprises stacked linguistic and acoustic components, with the two parts connected by upsampling the hidden states from linguistic part's top layer and feeding them into the acoustic component. In contrast to previous two-pass speech-to-speech models \citep{pmlr-v162-jia22b,inaguma-etal-2023-unity}, NAST-S2$x$ leverages its fully non-autoregressive nature.  It no longer relies on intermediate text decoding to determine the information passed to the acoustic component.
This characteristic allows it to be trained and tested directly from speech to acoustic units, thereby circumventing issues related to error propagation.

\subsubsection{Streaming Acoustic Encoder}
The acoustic encoder operates by setting a chunk size $T_s$. We extract FBank features from the streaming speech every $T_s$ ms, which are then fed into the encoder. The acoustic encoder consists of two layers of causal convolution for downsampling and followed by multiple standard Transformer layers. In a Transformer layer, features within each chunk are encoded bidirectionally, and the information from all previous chunks can also be attended to. Given the strong local dependencies in speech, we additionally employ Lookahead encoding \citep{liu-etal-2021-cross}, which enables states in each chunk to attend to its subsequent $r$ frames.

\subsubsection{Streaming Non-autoregressive Decoder}
Once the latest chunk is encoded, we use the features as input to the linguistic decoder. Given the significant discrepancy in length between the sequences of FBank and text, we downsample the encoded features before feeding them into the decoder:
\begin{equation}
\label{eq:downsample}
    \mathrm{DownSample}(\mathbf{\tilde{s}}^{e}_{i},r_{\mathrm{down}}),
\end{equation}
where $\mathbf{\tilde{s}}^{e}_{i}$ represents the encoded features in the $i$-th chunk and $r_{\mathrm{down}}$ is the downsampling ratio. We use $\mathrm{MeanPooling}$ applied to every $r_{\mathrm{down}}$ encoded features in our experiments.

The linguistic decoder also works in a chunk-by-chunk manner. The decoding of current chunk relies solely on hidden states in the previous chunks rather than any generated token:
\begin{equation}
\begin{aligned}
        &\mathrm{SelfAttn}(\mathbf{s}^{ld}_{i},\mathbf{s}^{ld}_{\leq i}),\\
        &\mathrm{CrossAttn}(\mathbf{s}^{ld}_{i},\mathbf{\tilde{s}}^{e}_{\leq i}),
\end{aligned}
\end{equation}
where $\mathbf{s}^{ld}_{i}$ denotes the hidden states in the $i$-th chunk in the linguistic decoder. Optionally, the linguistic decoder can generate text translation from the chunks. The text logits are derived by projecting the last layer states.

Meanwhile, hidden states in the last layer of linguistic decoder serve as input to the acoustic decoder after upsampling. This upsampling is designed to bridge the length gap between the sequences of text and acoustic unit:
\begin{equation}
\label{eq:upsample}
    \mathrm{UpSample}(\mathbf{\tilde{s}}^{ld}_{i},r_{\mathrm{up}}),
\end{equation}
where $\mathbf{\tilde{s}}^{ld}_{i}$ denotes the last layer states of the linguistic decoder in the $i$-th chunk and $r_{\mathrm{up}}$ is the upsampling ratio. We simply copy each state in the chunk $r_{\mathrm{up}}$ times. 

The acoustic decoder operates similarly to the linguistic decoder. Compared with previous two-pass models, our non-autoregressive acoustic decoder can directly attend to the acoustic encoder. This capability enables it to incorporate a broader range of acoustic information (e.g., rhythm, pitch, and energy) and helps in recovering from potential mistakes made by the linguistic decoder:
\begin{equation}
\centering
\begin{aligned}
        &\mathrm{SelfAttn}(\mathbf{s}^{ad}_{i},\mathbf{s}^{ad}_{\leq i}),\\
        &\mathrm{CrossAttn}(\mathbf{s}^{ad}_{i},\mathbf{\tilde{s}}^{e}_{\leq i}),
\end{aligned}
\end{equation}
where $\mathbf{s}^{ad}_{i}$ denotes the hidden states in the $i$-th chunk in the acoustic decoder. We use the states in the top layer to predict acoustic units.

When predicting text and unit sequences, an additional blank token is included in the vocabulary. The model dynamically adjusts the output length of each chunk by generating repeated or blank tokens. Subsequently, the collapse function in CTC \citep{10.1145/1143844.1143891} is employed for online deduplication and removal of blanks to generate the final output. The generated chunk of units is sent directly to a separate unit-based HiFi-GAN vocoder \citep{polyak21_interspeech} for synthesizing the waveform, which is then played immediately to the listener.

\subsection{Latency Control}
In this subsection, we explore various techniques for controlling the latency of NAST-S2$x$.

\hspace*{\fill} \\
\textbf{Chunk Size}\ Given that NAST-S2$x$ operates at a chunk level, a straightforward approach to controlling latency is to adjust the chunk size. Specifically, when the chunk size exceeds the utterance length, our model transitions into an offline model, conducting bidirectional encoding and bidirectional non-autoregressive decoding.

\hspace*{\fill} \\
\textbf{Lookahead}\ Chunk lookahead decoding resembles Lookahead encoding. When a feature chunk is sent to the decoder, it is allowed to wait for its subsequent $k$ chunks before starting decoding:
\begin{equation}
\begin{aligned}
    &\mathrm{CrossAttn}(\mathbf{s}^{ld}_{i},\mathbf{\tilde{s}}^{e}_{\leq i+k}),\\
    &\mathrm{CrossAttn}(\mathbf{s}^{ad}_{i},\mathbf{\tilde{s}}^{e}_{\leq i+k}).
\end{aligned}
\end{equation}
This allows the model to obtain more source-side information through an additional delay of $(k \cdot T_s)$ ms, without changing the chunk size.

\subsection{Training}
While NAST-S2$x$ benefits from various advantages of non-autoregressive generation, training it is challenging. Previous studies \citep{pmlr-v162-huang22k,NEURIPS2023_1c3d419b} have highlighted that NAR generation struggles to capture multi-modal distributions.
Regrettably, speech-to-speech translation encounters this multimodality problem.
This challenge stems from two aspects: First, the mapping from speech input to text translation can be one-to-many, as different word choices and grammar structures may convey the same semantics. Secondly, the distribution of speech when the text is given can be multi-modal, with variations in pitch, rhythm, and energy. To mitigate these challenges, we propose the following strategies to train NAST-S2$x$.

\subsubsection{Multi-task Non-monotonic Training}
Due to the performance decline observed in NAR models when trained with maximum likelihood estimation, we train NAST-S2$x$ using CTC-based non-monotonic latent alignment loss \citep{NEURIPS2022_35f805e6} 
\begin{equation}
    \mathcal{L}_{o}(\theta)=-\frac{2\cdot\sum_{g\in G_2} \min\{C_g(\bm{o}),C_g(\theta)\}}{\sum_{g\in G_2}(C_g(\bm{o})+C_g(\theta))},
\end{equation}
where $\bm{o} \in \{\bm{y}, \bm{z}\}$ is the target for either S2T or S2U task. $C_g(y)$ denotes the occurrence count of bigram $g$ in target, $C_g(\theta)$ represents the expected count of $g$ for model, and $G_2$ denotes the set of all bigrams in target. This training objective maximizes the F1 score of expected bigram matching between target and the uncollapsed output, and guides NAST-S2$x$ towards convergence to a concentrated distribution, thereby alleviating the multimodality problem in speech-to-speech translation. We utilize multi-task learning to integrate the losses from both text and acoustic unit prediction tasks into our training process:
\begin{equation}
    \mathcal{L}=\mathcal{L}_{\bm{y}}(\theta)+\mathcal{L}_{\bm{z}}(\theta).
\end{equation}

\subsubsection{Two-Step Glancing}
To further simplify the learning complexity for both linguistic and acoustic decoders, we further introduce the concept of glancing \citep{qian-etal-2021-glancing} to our NAST-S2$x$ training.
As depicted in Figure \ref{architecture}, we find the most probable sequence that can be collapsed to the target within the current distribution of text and acoustic unit in the model:
\begin{equation}
\begin{aligned}
&\bm{a}_{\mathrm{unit}}^{*} = \mathop{\arg \max}\limits_{\bm{a}_{\mathrm{unit}} \in \beta^{-1}(\bm{z})} {p_{\theta}(\bm{a}_{\mathrm{unit}}|\bm{x})},\\
&\bm{a}_{\mathrm{text}}^{*} = \mathop{\arg \max}\limits_{\bm{a}_{\mathrm{text}} \in \beta^{-1}(\bm{y})} {p_{\theta}(\bm{a}_{\mathrm{text}}|\bm{x})},    
\end{aligned}
\end{equation}
 where $\bm{a}_{\mathrm{unit}}$ and $\bm{a}_{\mathrm{text}}$ represent the predicted uncollapsed sequence of text and acoustic unit, and $\beta^{-1}$ is the inverse of collapse function. We then randomly substitute the features fed to both the linguistic and acoustic decoders with token embeddings corresponding to positions in the most probable text or unit sequences.
 
 This strategy simplifies the complexity of S2S mapping by providing cues during both decoding stages. This induces the NAR model to learn a deterministic conditional distribution, mitigating the issue of insufficient capacity for tasks with multi-modal distributions.

 \section{Experiments}
\subsection{Speech-to-Text}
\textbf{Datasets} We conduct experiments on two MuST-C\footnote{\url{https://ict.fbk.eu/must-c}} language pairs: English to German (En$\rightarrow$De) and English to Spanish (En$\rightarrow$Es) \citep{di-gangi-etal-2019-must}. We use the \texttt{dev} set for validation and report performance on the \texttt{tst-COMMON} set.

\hspace*{\fill} \\
\textbf{Pre-processing}
The input speech is represented as 80-dimensional log mel-filterbank coefficients computed every 10ms with a 25ms window. Global channel mean and variance normalization is applied to the input speech. In training, SpecAugment \citep{park2019specaugment} data augmentation with the LB policy is additionally employed. We use SentencePiece \citep{kudo-richardson-2018-sentencepiece} to generate a unigram vocabulary of size 10000 for the source and target text jointly.

\hspace*{\fill} \\
\textbf{Model Configurations}
In the Simul-S2T experiments, we exclusively utilize the linguistic component of the decoder. We set the downsampling ratio\footnote{For details on the analysis of the downsampling ratio, see Appendix \ref{app:analysis_ratio}.} to 2 and explore chunk sizes within the set $\{160, 320, 640, 1280\}$ ms. The offline results are obtained by setting the chunk size to be longer than any utterance in the corpus. The number of additional frames the encoder can attend to is set equal to the size of a chunk. When employing lookahead decoding, we vary the lookahead number $k$ within the range $\{0, 2, 6\}$ while maintaining a fixed chunk size of 320ms. More implementation details can be found in Appendix \ref{app:implementation details}.

\hspace*{\fill} \\
\textbf{Baselines} We compare our NAST-S2$T$ with several strong Simul-S2T baselines. Further details regarding baselines are available in Appendix \ref{app:baselines_s2t}.

\hspace*{\fill} \\
\textbf{Evaluation} We use SimulEval\footnote{\url{https://github.com/facebookresearch/SimulEval}} toolkit \citep{ma-etal-2020-simuleval} for evaluation. Translation quality is assessed using case-sensitive detokenized \texttt{BLEU} \citep{papineni-etal-2002-bleu, post-2018-call}, while latency is measured by word-level Average Lagging \citep[\texttt{AL};][]{ma-etal-2020-simulmt}. Numerical results with more latency metrics are provided in Appendix \ref{app:numerical_s2t}.

\subsubsection{Preliminary Experiment}
We first conduct a preliminary experiment to compare latency control strategies. Employing NAST-S2$T$ with a baseline chunk size of 320ms, we examine the trade-off between latency and quality by adjusting the chunk size and implementing lookahead decoding. As depicted in Table \ref{table:strategy}, both stratigies enhance quality at the sacrifice of latency. Nevertheless, increasing the chunk size yields superior quality with reduced latency over lookahead decoding. Notably, there appears to be a quality plateau when utilizing lookahead decoding. Waiting for an extra 6 source chunks versus 2 extra ones results in nearly identical quality, despite an additional delay of almost 1000ms. This implies that the amount of source information alone does not solely dictate translation quality. By adopting the strategy of increasing chunk size, we not only enable the model to attend to more source information but also facilitate bidirectional non-autoregressive decoding of longer sequences within a chunk. This enhancement significantly improves the translation quality. Therefore, we only vary the chunk size in the main experiment.

\begin{table}[t]
\small
\centering
\begin{tabular}{cc|ccc}
\toprule
  \multirow{3}{*}{\textbf{Chunk}}  &  \textbf{$ms$}            &  $\mathbf{320}$  & $\mathbf{640}$ & $\mathbf{1280}$ \\ 

  & \textbf{BLEU}  & 26.48 & 27.02 & 28.05  \\
 ($\#=0$) & \textbf{AL}  & 1114 & 1396 & 2180  \\
 \midrule
  \multirow{3}{*}{\textbf{Lookahead}}  &  \textbf{$\#$}            &  $\mathbf{0}$  & $\mathbf{2}$ & $\mathbf{6}$ \\ 

  & \textbf{BLEU}  & 26.48 & 27.02 & 26.99  \\
 ($320ms$) & \textbf{AL}  & 1114 & 1762 & 2781  \\
 
\bottomrule
\end{tabular}
\caption{Results of the quality-latency trade-off with increasing the chunk size or implementing lookahead decoding. Experiments are conducted on En$\rightarrow$Es Simul-S2T task.}
\label{table:strategy}
\end{table}

\begin{figure*}[ht]
\centering
\subfigure[En$\rightarrow$De]{
\includegraphics[width=2.6in]{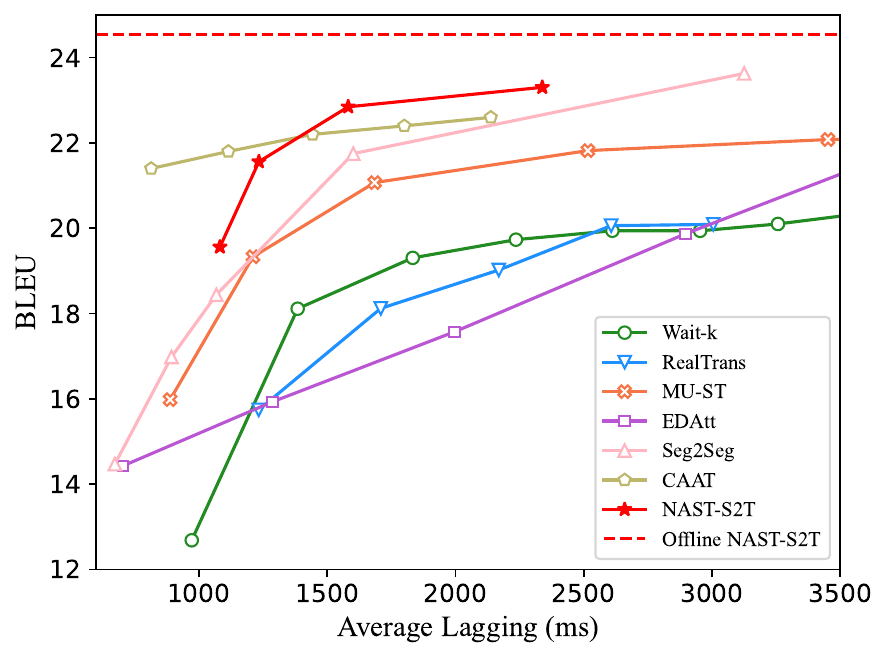}
}\hspace{2.6mm}
\subfigure[En$\rightarrow$Es]{
\includegraphics[width=2.6in]{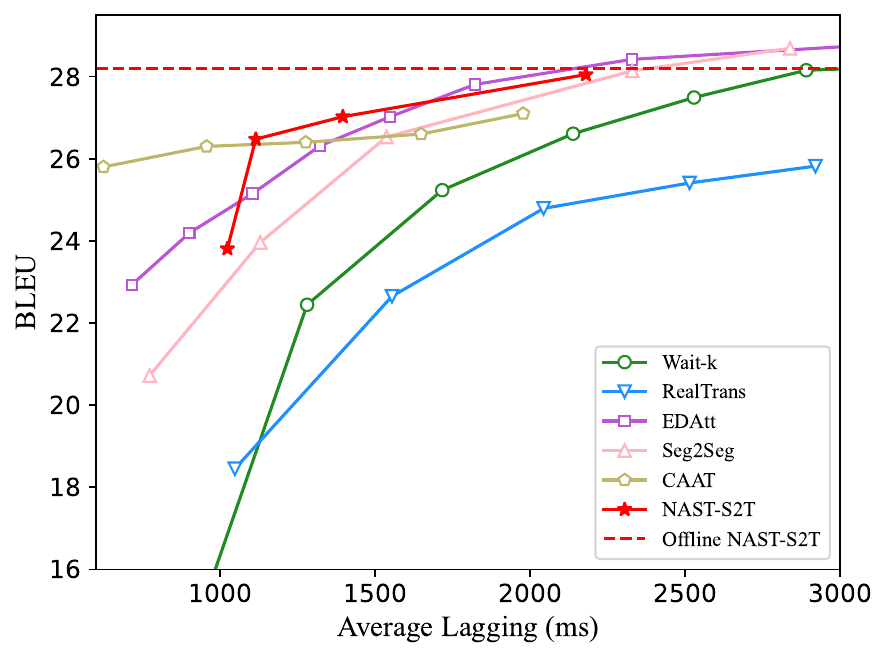}
}

\caption{Results of translation quality (\texttt{BLEU}) against latency (Average Lagging, \texttt{AL}) on MuST-C En$\rightarrow$De and En$\rightarrow$Es datasets. The red solid line and dashed line illustrate the performance of NAST-S2$T$ under different chunk sizes $T_s$ or in an offline condition. The numerical results are presented in Table \ref{table:ende} and Table \ref{table:enes}.}
\vspace{-2mm}
\label{main_s2t}
\end{figure*}

\subsubsection{Main Results and Analysis}
Figure \ref{main_s2t} illustrates the main results of Simul-S2T task. Detailed numerical results are available in Table \ref{table:ende} and \ref{table:enes}. It can be observed that NAST-S2$T$ achieves competitive or superior translation quality compared to strong baselines across various latency constraints. At lower latency, its performance is only inferior to CAAT \citep{liu-etal-2021-cross}. Meanwhile, it performs better or comparably as the autoregressive models under higher latency or offline conditions.
Both datasets demonstrate that as the chunk size $T_s$ increases from 160ms to 320ms, there is a significant improvement in translation quality with only a minor increase in latency. We attribute this phenomenon to the average duration of each word, estimated to be approximately 280ms \citep{ma-etal-2020-simulmt}. The model's performance tends to degrade when the chunk size falls below it. Furthermore, we find that NAST-S2$T$ achieves a better balance when the chunk size $T_s$ is 640ms (\texttt{AL} $\approx$ 1200ms), after which the quality gain from further increasing the chunk size diminishes.

\subsection{Speech-to-Speech}
\label{sec:exp:s2s}

\textbf{Datasets} We conduct experiments on CVSS-C\footnote{\url{https://github.com/google-research-datasets/cvss}} French to English (Fr$\rightarrow$En) dataset \citep{jia2022cvss}. 

\hspace*{\fill} \\
\textbf{Pre-processing} For the source speech, we resample the audio to 16kHz and apply identical preprocessing steps as those used in the Simul-S2T experiments. For the target speech, we also downsample the audio and extract discrete units utilizing the publicly available pre-trained mHuBERT model and K-means quantizer.\footnote{\url{https://github.com/facebookresearch/fairseq/blob/main/examples/speech_to_speech/docs/textless_s2st_real_data.md}}

\hspace*{\fill} \\
\textbf{Model Configurations}
The downsampling and upsampling ratio are set to 2 and 6. We explore different settings for chunk sizes within the set $\{320, 640, 1280, 1920, 2560\}$ ms. The offline results are obtained by setting the chunk size to be longer than any utterance. The number of additional frames the encoder can attend to is set equal to the size of a chunk. We also experimented with fixing the duration of additional frames to 1280ms when the chunk size is larger.
More details can be found in Appendix \ref{app:implementation details}.

\hspace*{\fill} \\
\textbf{Baselines} 

\textbf{Wait-$k$-Stride-$n$}: We employ Wait-$k$ strategy \citep{ma-etal-2019-stacl} for S2UT model \citep{{lee-etal-2022-direct}} to build an end-to-end Simul-S2S baseline. Since the input is speech audio, a pre-decision module is needed to segment the utterance into multiple chunks to execute Wait-$k$ \citep{ma-etal-2020-simulmt}. Furthermore, the translation of a speech chunk can consist of multiple acoustic units to form the pronunciation of a word. It is reasonable to generate multiple unit tokens upon receiving a speech chunk. Therefore, we adopt Wait-$k$-Stride-$n$ strategy \citep{zeng-etal-2021-realtrans} to construct an end-to-end Simul-S2S baseline, varying the speech chunk size and the hyperparameters $k$ and $n$. The numerical results can be found in Table \ref{table:fren_waitk}.

\textbf{EDAtt + Tacotron2}: We further provide the results of cascade systems (Simul-S2T + TTS) for comparison. We choose EDAtt \citep{papi-etal-2023-attention} as the Simul-S2T model. According to the recommendation in \citet{papi-etal-2023-attention}, we train a Conformer + CTC compression model \citep{gaido-etal-2021-ctc} with a total of
$\sim$120M parameters using speech-text parallel pairs of CVSS-C Fr-En dataset as the offline model to implement EDAtt algorithm. For TTS part, we use a Tacotron2 model trained on LJSpeech\footnote{\url{https://huggingface.co/speechbrain/tts-tacotron2-ljspeech}}. Whenever the Simul-S2T model generates a complete word, we send it to the TTS model and generate a speech chunk as output. The numerical results can be found in Table \ref{table:fren_edatt}.

We also compare NAST-S2$S$ with several strong Offline-S2S models to assess its performance in offline scenarios. Further details regarding baselines are available in Appendix \ref{app:baselines_s2s}.

\hspace*{\fill} \\
\textbf{Evaluation} We also use SimulEval toolkit for evaluation. Following \citet{ma2022direct}, we keep discontinuities between generated speech chunks to simulate real-world scenarios. Translation quality is assessed using \texttt{ASR-BLEU}. We also employ \texttt{BLASER 2.0}\footnote{\url{https://huggingface.co/facebook/blaser-2.0-ref}} \citep{seamlessm4t2023} to assess the quality. The results for \texttt{BLASER 2.0} are presented in Table \ref{table:fren_blaser2}.
Regarding latency, we report \texttt{AL} and \texttt{AL\_EOW} \citep{ma2022direct}. \texttt{AL} measures time delay of waveform chunks, while \texttt{AL\_EOW} assesses the delay of text transcribed from generated speech. 
The generated time of each word is considered as the end time of its corresponding segment. 
Numerical results with more latency metrics are provided in Appendix \ref{app:numerical_s2s}.

\begin{table*}[ht]
\small
\centering
\begin{tabular}{c|c|c|c|c|c} 
\toprule
 \textbf{Model} & \textbf{\#Params} & \textbf{End-to-End} & \textbf{Streamable} & \textbf{ASR-BLEU} &  \textbf{Speedup} \\
 \midrule
S2UT \scriptsize{\citep{lee-etal-2022-direct}}   &   58M        &   \usym{2713}     &  \usym{2717}    &   24.80   &  1.00× \\      
UnitY \scriptsize{\citep{inaguma-etal-2023-unity}} &  67M        &   \usym{2717}     &   \usym{2717}   &  26.90    &  1.60× \\    
DASpeech \scriptsize{\citep{NEURIPS2023_e5b1c0d4}} &   93M        &  \usym{2717}     &   \usym{2717}    &   25.03   &  16.29× \\ 
Offline NAST-S2S &   79M        &   \usym{2713}   &  \usym{2713}      &   25.82   &  28.30× \\ 
 \bottomrule
\end{tabular}
\caption{Comparison of strong Offline-S2S baselines and our NAST-S2$S$ in offline conditions. The speedup is measured using a GeForce RTX 3090 GPU with a batch size of 1.}
\label{table:offline_nat}
\end{table*}

\subsubsection{Main Results}

\begin{figure}[htbp]
\centering
\includegraphics[width=2.6in]{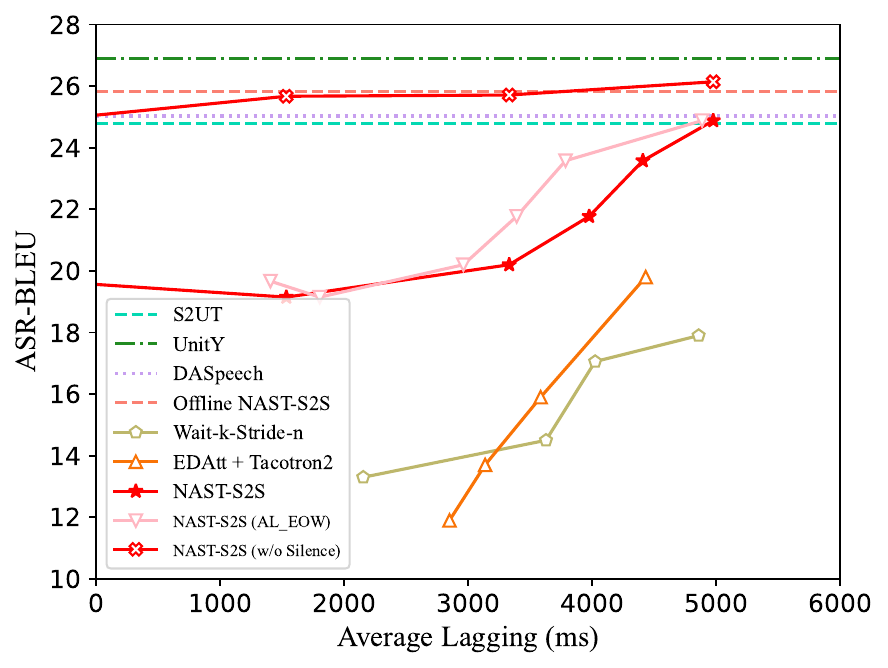}
\caption{Results of translation quality in offline conditions and simultaneous scenarios (\texttt{ASR-BLEU} or \texttt{ASR-BLEU (Silence Removed)} against \texttt{AL} or \texttt{AL\_EOW}). The numerical results of NAST-S2$S$ are presented in Table \ref{table:fren} and Table \ref{table:fren_asr_bleu_rm}.}
\label{main_s2s}
\end{figure}

Figure \ref{main_s2s} illustrates the main results of Simul-S2S task. Detailed numerical results are presented in Table \ref{table:fren}. We observe a trend where the translation quality of NAST-S2$S$ generally improves as latency increases, with a notable improvement from 3000ms to 4000ms. Even under extremely low latency conditions (\texttt{AL} $\approx$ 1000ms), NAST-S2$S$ still achieves acceptable translation quality (\texttt{ASR-BLEU} > 19). This result even surpasses the performance of wait-$k$-stride-$n$ and cascade baselines at 4000ms latency. 
Furthermore, we discover that in offline scenarios, the quality achieved by NAST-S2$S$ exceeds that of the current leading NAR Offline-S2S model DASpeech \citep{NEURIPS2023_e5b1c0d4} by nearly 1 \texttt{ASR-BLEU}, with translation quality only slightly inferior to two-pass autoregressive model UnitY\footnote{Two-pass models are not strictly end-to-end, as they must generate target text before producing the speech output.} \citep{inaguma-etal-2023-unity}.
\subsubsection{Analysis on Inference Efficiency}
\begin{table*}[ht]
\small
\centering
\begin{tabular}{c|c|ccc|ccc|ccc|c} 
\toprule
 \multirow{2}{*}{\textbf{$T_s$ \scriptsize{($ms$)}}} & \multirow{2}{*}{\textbf{ASR-BLEU}} & \multicolumn{3}{c}{\textbf{Average Lagging \scriptsize{($ms$)}}} & \multicolumn{3}{c}{\textbf{Start Offset \scriptsize{($ms$)}}} & \multicolumn{3}{c}{\textbf{End Offset \scriptsize{($ms$)}}} &  \multirow{2}{*}{\textbf{ACT \scriptsize{($ms$)}}}  \\
 & & NCA & CA & $\Delta$ & NCA & CA & $\Delta$ & NCA & CA & $\Delta$ & \\

 \midrule
 $\mathbf{320}$ & 19.67 & -392 & 347 & 739 & 655 & 712 & 57 & 562 & 1550 & 988 & 555\\
 $\mathbf{640}$ & 19.15 & 1532 & 1824 & 292 & 1294 & 1350 & 56 & 863 & 1344 & 481 & 297\\
 $\mathbf{1280}$ & 20.20 & 3330 & 3500 & 170  & 2566 & 2642 & 76 & 1648  & 1901& 253 & 192\\
 $\mathbf{2560}$ & 24.88 & 4975 & 5097 & 122 & 4691 & 4781 & 90 & 2753  & 2879& 126 & 120\\
 \bottomrule
\end{tabular}
\caption{Results of translation quality (\texttt{ASR-BLEU}), latency (\texttt{Average Lagging}, \texttt{Start Offset} \& \texttt{End Offset}) and average computation time per chunk generation (\texttt{ACT}) during NAST-S2$S$ simultaneous inference. All latency metrics report both the computation-aware (\texttt{CA}) version and the non-computation-aware (\texttt{NCA}) version, as well as their differences (\texttt{$\Delta$}).} 
\label{table:simul_nat}
\end{table*}
Speech-to-speech translation imposes strong demands on inference efficiency. In Offline-S2S, efficiently generating long sequences of acoustic unit is crucial to minimize waiting time. In Simul-S2S, reducing computational time overhead is essential to avoid extra latency.
Benefiting from end-to-end non-autoregressive generation, NAST-S2$S$ offers appealing advantages in both scenarios. Table \ref{table:offline_nat} presents the comparison in Offline-S2S. NAST-S2$S$ achieves a 28× speedup compared to S2UT and a 17× speedup compared to UnitY at decoding. In Simul-S2S, the advantage in inference speed becomes more critical. Table \ref{table:simul_nat} presents the comparison of non-computation-aware and computation-aware latency. The gap between \texttt{AL} and \texttt{AL\_CA} and the average computation time per chunk generation are both less than 300ms when the chunk size is larger than 640ms, indicating that NAST-S2$S$'s latency in practical use is similar to the theoretical latency of its simultaneous translation policy.

\subsubsection{Analysis on Discontinuity}
We observed notable differences in the performance of NAST-S2$x$ between Simul-S2S and Simul-S2T tasks. NAST-S2$T$ achieves satisfactory quality when the chunk size $T_s$ is set to 640ms (\texttt{AL} < 2000ms). However, to attain translation quality comparable to offline condition, NAST-S2$S$ requires an increase in the chunk size $T_s$ to 2560ms. This discrepancy may stem from the differing nature of text and speech streaming generation. In text generation, appending newly generated chunk directly after the historical sequence is straightforward. However, in speech generation, there may be silence intervals between each speech chunk, particularly when the chunk size $T_s$ exceeds the duration of the last generated speech chunk. Therefore, we speculate that as the chunk size decreases, increased silence between generated speech chunks may lead to discontinuity in speech, thereby decreasing the overall quality.

To validate this hypothesis, we further analyze the trends of the following metrics as the chunk size varies: \texttt{ASR-BLEU} \texttt{(Silence Removed)}, representing \texttt{ASR-BLEU} score after removing the added silence between generated chunk; \texttt{Unit-BLEU}, representing \texttt{BLEU} score of the generated unit sequences against the reference; \texttt{S2T-BLEU}, where we conduct additional decoding of the linguistic decoder to evaluate quality in Simul-S2T.
We also provide statistics on the number of discontinuities (\texttt{DCNum}), the average silence duration per discontinuity (\texttt{DCAve}), and the total silence duration (\texttt{DCSum}) in the generated streaming speech.
\begin{table}[t]
\small
\centering
\begin{tabular}{c|cccc} 
\toprule
 $T_s$\ \scriptsize{$(ms)$} & $\mathbf{320}$ & $\mathbf{640}$ & $\mathbf{1280}$ &  $\mathbf{2560}$ \\
 \midrule
  \textbf{S2T-BLEU} & 28.04 & 28.28 & 28.23 &  28.78 \\
  \textbf{Unit-BLEU} & 33.41 & 33.97 & 34.04 &  34.40 \\
 \textbf{ASR-BLEU} & 19.67 & 19.15 & 20.20 &  24.88 \\
  \textbf{ASR-BLEU} & \multirow{2}{*}{24.90} & \multirow{2}{*}{25.67} & \multirow{2}{*}{25.71} &  \multirow{2}{*}{26.14} \\
  (\texttt{Silence Removed}) & & & & \\
 \midrule
 \textbf{DCNum} & 7.3 & 4.7 & 2.1 &  0.4 \\
 \textbf{DCAve \scriptsize{$(ms)$}} & 355 & 450 & 685 &  360 \\
 \textbf{DCSum \scriptsize{$(ms)$}} & 2220 & 1952 & 1420 &  395 \\
 \bottomrule
\end{tabular}
\caption{Statistics of NAST-S2$S$ generation across varying chunk sizes $T_s$.}
\label{table:statistics}
\end{table}

Table \ref{table:statistics} presents the statistics. We observed minor degradation in the values of \texttt{Unit-BLEU} and \texttt{S2T-BLEU} even at a chunk size of 320ms, showing NAST-S2$S$'s capability in streaming text and unit generation. However, there exists a significant increase in the number of discontinuities as the chunk size decreases. Although the duration of silence per discontinuity is relatively short when the chunk size is small, the increase in their number results in a longer total silence duration, thus intensifying the degree of discontinuity and impacting its overall quality (\texttt{ASR-BLEU}). 

Moreover, if the added silence were removed, the measured \texttt{ASR-BLEU} \texttt{(Silence Removed)} significantly increased and the gap between streaming and offline scenarios becomes small. This suggests that \texttt{ASR-BLEU} may underestimate speech quality here. The decline in \texttt{ASR-BLEU} scores is primarily due to the \emph{playback timing}. For example, consider the word "\emph{Richardson}", which consists of multiple syllables. If the "\emph{Richard}" part of the waveform is generated in the previous chunk and played immediately, and the "\emph{son}" syllable is generated in the subsequent chunk, the potential silence period (which equals to the chunk size minus the length of the waveform generated in the previous chunk) could cause the listener to perceive a \emph{stuttering} effect, leading to a decrease in \texttt{ASR-BLEU} scores.

\section{Related Work}
Researches in simultaneous speech translation can be roughly categorized into Simul-S2T \citep{ma-etal-2020-simulmt} and Simul-S2S \citep{zheng-etal-2020-fluent} variants.

\hspace*{\fill} \\
\textbf{Simul-S2T}
With the rise of neural networks, Simul-S2T models no longer rely on the transcription as a bridge \citep{ma-etal-2020-simulmt,iranzo-sanchez-etal-2020-direct}. Given the difference between speech and text input, some researchers focus on how to divide speech chunks and then execute strategies. \citet{ma-etal-2020-simulmt} employed fixed-length segmentation and implemented Wait-$k$ \citep{ma-etal-2019-stacl} and MMA \citep{Ma2020Monotonic} based on that; \citet{ren-etal-2020-simulspeech,zeng-etal-2021-realtrans,chen-etal-2021-direct} utilized ASR results to partition and execute Wait-$k$ or its variants. \citet{zhang-etal-2022-learning} trained a segmentation model to detect semantic units. \citet{zhang-feng-2023-end} trained a model to dynamically segment with differentiable approach, then extending it to a segment-to-segment framework \citep{NEURIPS2023_8df70595}. Additionally, some researchers have also attempted to use Transducer \citep{DBLP:journals/corr/abs-1211-3711} and incorporate attention mechanisms to enhance its performance \citep{liu-etal-2021-cross,tang-etal-2023-hybrid}.
Besides, some researchers are leveraging offline models for simultaneous inference. \citet{liu20s_interspeech} considered the agreeing prefixes of two consecutive chunks as stable hypotheses. 
\citet{papi-etal-2023-attention,papi23_interspeech} used attention as guidance, allowing the model to generate output for the current step if its attention is not focused on the most recently received frames. 

\hspace*{\fill} \\
\textbf{Simul-S2S}
There have been limited prior studies exploring Simul-S2S. \citet{zheng-etal-2020-fluent} and \citet{sudoh2020simultaneous} both developed cascade models by integrating streaming ASR, Simul-T2T, and incremental TTS components. Additionally, \citet{Liu2021FromST} proposed latency reduction strategies for incremental TTS in Simul-S2S. Moreover, \citet{ma2022direct} introduced a variational version of MMA to S2UT \citep{lee-etal-2022-direct} and constructed the first end-to-end Simul-S2S model.
\section{Conclusion}
In this paper, we present a non-autoregressive streaming generation framework for simultaneous speech-to-any translation, which integrates both Simul-S2T and Simul-S2S tasks into a unified framework. Experimental results on various benchmarks showcase the superiority of our model.

\section*{Limitation}
Our NAST-S2$x$ exhibits greater latency in Simul-S2S compared to Simul-S2T tasks. This discrepancy arises due to NAST-S2$S$'s reliance on an external vocoder, typically trained on offline tasks and not adapted for streaming scenarios, thereby constraining NAST-S2$S$'s performance.
Additionally, our method requires a parallel speech-to-speech translation corpus for end-to-end training, which can be challenging to obtain. Existing datasets are typically based on synthesized target speech. The lack of such corpora may hinder the development of simultaneous speech-to-speech translation models.

\section*{Acknowledgement}
We thank the anonymous reviewers for their insightful comments. This work is supported by National Natural Science Foundation of China (Grant No. 62376260).

\bibliography{custom}

\clearpage

\appendix
\section{Implementation Details}
\label{app:implementation details}
\subsection{Configuration}
We incorporate both cosine positional encoding \citep{NIPS2017_3f5ee243} and relative positional attention \citep{shaw-etal-2018-self} into the acoustic encoder, and utilize learned positional encoding for non-autoregressive decoder. A separate learned positional encoding is applied to the acoustic decoder. The acoustic encoder comprises two layers of causal convolution followed by six standard Transformer layers. Both the non-autoregressive linguistic and acoustic decoders consist of six Transformer layers each. All Transformer layers are configured with a 512 embedding dimension, 8 attention heads, and a 2048 FFN dimension.
The total number of parameters for NAST-S2$T$ and NAST-S2$S$ are 52M and 79M.
\subsection{Training}
\textbf{NAST-S2$T$}\ Considering the inherent complexity of speech-to-text translation, we leverage the concept of curriculum learning. We initialize the encoder of NAST-S2$T$ with an ASR-trained model and conduct pretraining using CTC loss \citep{10.1145/1143844.1143891}. Subsequently, we employ non-monotonic training to further refine NAST-S2$T$. During the CTC loss pretraining, we set the dropout rate to $0.3$, weight decay to $0.01$, and incorporate label smoothing with a value of $0.01$. The dropout rates for activation and attention are both set to $0.1$. The pretraining process spans $100k$ updates with a batch size of $320k$ tokens. The learning rate gradually warms up to $1 \cdot 10^{-3}$ within $10k$ steps, while the text glancing ratio linearly anneals from $0.5$ to $0.3$ over $50k$ steps. In non-monotonic training, we adjust the dropout rate to $0.1$ while keeping other hyperparameters unchanged. This stage involves training NAST-S2$T$ for $20k$ updates. The learning rate warms up to $3 \cdot 10^{-4}$ within $4k$ steps, and the text glancing ratio is maintained at $0.3$. Throughout the training, we optimize models using the Adam optimizer \citep{DBLP:journals/corr/KingmaB14} with parameters $\beta=(0.9,0.98)$ and $\epsilon = 10^{-8}$. We utilize sequence-level knowledge distillation \citep{kim-rush-2016-sequence} solely during the CTC pretraining stage to facilitate model warmup, while NAST-S2$T$ is trained directly on raw data during non-monotonic training.

\textbf{NAST-S2$S$}\ Similar to the training of NAST-S2$T$, a curriculum learning approach is also devised for NAST-S2$S$. We initialize the encoder of NAST-S2$S$ with an ASR-trained model and conduct multi-task pretraining using the CTC loss. Subsequently, we employ multi-task non-monotonic training to further refine NAST-S2$S$. During the pretraining, the hyperparameters are consistent with those used in NAST-S2$T$, with the exception of incorporating label smoothing for both text and unit targets, set at a value of $0.01$. The multi-task pretraining process spans $50k$ updates with a batch size of $320k$ tokens. The text glancing ratio linearly anneals from $0.5$ to $0.3$ over $50k$ steps, while the unit glancing ratio linearly decreases from $0.3$ to $0.1$ over the same number of steps. In multi-task non-monotonic training, we adjust the dropout rate to $0.1$ while keeping other hyperparameters unchanged. This stage involves training NAST-S2$S$ for $30k$ updates. The learning rate warms up to $3 \cdot 10^{-4}$ within $4k$ steps. We maintain a text glancing ratio of $0.3$ and a unit glancing ratio of $0.1$ in this stage. Knowledge distillation is not utilized during the entire training of NAST-S2$S$.

\section{Baselines}
\label{app:baselines}
\subsection{Speech-to-Text}
\label{app:baselines_s2t}
We compare our NAST-S2$T$ with the following strong Simul-S2T baselines. 

\textbf{Wait-$k$} \citep{ma-etal-2020-simulmt}: It executes Wait-$k$ policy \citep{ma-etal-2019-stacl} by setting the pre-decision window size to 280 ms.

\textbf{RealTrans} \citep{zeng-etal-2021-realtrans}: It detects word number in the streaming speech by counting blanks in CTC transcription and applies Wait-$k$-Stride-$n$ strategy accordingly.

\textbf{MU-ST} \citep{zhang-etal-2022-learning}: 
It trains an external segmentation model, which is then utilized to detect meaningful units for guiding generation.

\textbf{Seg2Seg} \citep{NEURIPS2023_8df70595}: It alternates between waiting for a source segment and generating a target segment in an autoregressive manner.

\textbf{CAAT} \citep{liu-etal-2021-cross}: It utilizes the Transformer Transducer \cite{DBLP:journals/corr/abs-1211-3711, 9053896} as its foundational architecture for streaming generation and incorporates a cross-attention mechanism within the joiner module to alleviate the strong monotonic constraint.

\textbf{EDAtt} \citep{papi-etal-2023-attention}: It computes the attention scores towards the latest received speech frames, serving as guidance for an offline-trained speech translation model during simultaneous inference. The experimental results reported in their paper were obtained using a 112M parameter Conformer \citep{gulati20_interspeech,papi2023good}. To ensure a fair comparison with our method, we retrained a Conformer\footnote{\url{https://github.com/hlt-mt/FBK-fairseq/blob/master/fbk_works/BUGFREE_CONFORMER.md}} of similar size to NAST-S2$T$ on the same dataset to perform EDAtt decoding (52M parameters, achieved by reducing the encoder embedding dimension from 512 to 256 and keeping the number of encoder layers at 12). The numerical results of our re-implemented EDAtt can be found in Tables \ref{table:ende_edatt} and \ref{table:enes_edatt}.

\subsection{Speech-to-Speech}
\label{app:baselines_s2s}
We compare our NAST-S2$S$ with several strong Offline-S2S and Simul-S2S baselines.\\
\textbf{Offline-S2S}

\textbf{S2UT} \citep{lee-etal-2022-direct}: A direct speech-to-unit model, which predicts acoustic units in a standard autoregressive manner.

\textbf{UnitY} \citep{inaguma-etal-2023-unity}: A two-pass speech-to-unit model, which first generates a subword sequence in an autoregressive manner and then feeds the last hidden states into another autoregressive model to generate unit sequence.

\textbf{DASpeech} \citep{NEURIPS2023_e5b1c0d4}: A two-pass non-autoregressive speech-to-spectrogram model. It initially employs a directed acyclic graph layer \citep{huang2022DATransformer} to generate a phoneme sequence, followed by utilizing FastSpeech2 \citep{ren2021fastspeech} to synthesis the phonemes into mel-spectrograms.
\\
\textbf{Simul-S2S}

\textbf{Wait-$k$-Stride-$n$}: We employ Wait-$k$ strategy \citep{ma-etal-2019-stacl} for S2UT model \citep{{lee-etal-2022-direct}} to build an end-to-end Simul-S2S baseline. Since the input is speech audio, a pre-decision module is needed to segment the utterance into multiple chunks to execute Wait-$k$ \citep{ma-etal-2020-simulmt}. Furthermore, the translation of a speech chunk can consist of multiple acoustic units to form the pronunciation of a word. It is reasonable to generate multiple unit tokens upon receiving a speech chunk. Therefore, we adopt Wait-$k$-Stride-$n$ strategy \citep{zeng-etal-2021-realtrans} to construct an end-to-end Simul-S2S baseline, varying the speech chunk size and the hyperparameters $k$ and $n$. The numerical results can be found in Table \ref{table:fren_waitk}.

\textbf{EDAtt + Tacotron2}: We further provide the results of cascade systems (Simul-S2T + TTS) for comparison. We choose EDAtt \citep{papi-etal-2023-attention} as the Simul-S2T model. According to the recommendation in \citet{papi-etal-2023-attention}, we train a Conformer + CTC compression model \citep{gaido-etal-2021-ctc} with a total of $\sim$120M parameters using speech-text parallel pairs of CVSS-C Fr-En dataset as the offline model to implement EDAtt algorithm. For TTS part, we use a Tacotron2 model trained on LJSpeech. Whenever the Simul-S2T model generates a complete word, we send it to the TTS model and generate a speech chunk as output. The numerical results can be found in Table \ref{table:fren_edatt}.

\section{Numerical Results}
\label{app:numerical}
\subsection{Speech-to-Text}
\label{app:numerical_s2t}
In addition to Average Lagging \citep[\texttt{AL};][]{ma-etal-2020-simulmt}, we also incorporate Average Proportion \citep[\texttt{AP};][]{DBLP:journals/corr/ChoE16}, Differentiable Average Lagging \citep[\texttt{DAL};][]{arivazhagan-etal-2019-monotonic} and Length Adaptive Average Lagging \citep[\texttt{LAAL};][]{papi-etal-2022-generation} as metrics to evaluate the latency of NAST-S2$T$. \texttt{AL}, \texttt{DAL} and \texttt{LAAL} are reported with milliseconds. The trade-off between latency and quality is attained by adjusting the chunk size $T_s$. The offline results are obtained by setting the chunk size to be longer than any utterance in the dataset ($T_s = \infty$). We use SimulEval \texttt{v1.1.4} for evaluation in all the experiments. The numerical results of NAST-S2$T$ are presented in Table \ref{table:ende} and \ref{table:enes}.

\begin{table*}[h]
\centering
\begin{tabular}{c|cccc|c} \hline
\multicolumn{6}{c}{\textit{\textbf{NAST-S2T on En$\rightarrow$De}}}                                  \\\hline
 $T_s (ms)$ & \textbf{AP} & \textbf{AL} & \textbf{DAL} & \textbf{LAAL} & \textbf{BLEU} \\ \hline

160                  & 0.58        & 1082        & 1359        & 1191         & 19.51      \\
320                  & 0.65        & 1234        & 1546        & 1346         & 21.56          \\
640                  & 0.73        & 1582        & 1969        & 1692         & 22.85        \\
1280                 & 0.81        & 2338        & 2812        & 2423         & 23.30         \\
$\infty$                 & -        & -        & -        & -         & 24.54         \\

\hline
\end{tabular}
\caption{Numerical results of NAST-S2$T$ on MuST-C English to German speech-to-text translation dataset.}
\label{table:ende}
\end{table*}

\begin{table*}[h]
\centering
\begin{tabular}{c|cccc|c} \hline
\multicolumn{6}{c}{\textit{\textbf{NAST-S2T on En$\rightarrow$Es}}}                                  \\\hline
 $T_s (ms)$ & \textbf{AP} & \textbf{AL} & \textbf{DAL} & \textbf{LAAL} & \textbf{BLEU} \\ \hline

160                   &  0.62       &   1023      &   1541      &   1242       &   23.81       \\
320                  &   0.71       &   1114      &   1692    &    1377      &   26.48       \\
640                  &   0.79       &   1396      &    2030   &   1648       &   27.02       \\
1280                 &   0.86       &   2180      &    2843   &   2364       &   28.05       \\
$\infty$                 & -        & -        & -        & -         & 28.21         \\

\hline
\end{tabular}
\caption{Numerical results of NAST-S2$T$ on MuST-C English to Spanish speech-to-text translation dataset.}
\label{table:enes}
\end{table*}

\begin{table*}[h]
\centering
\begin{tabular}{c|cccc|c} \hline
\multicolumn{6}{c}{\textit{\textbf{EDAtt on En$\rightarrow$De}}}                                  \\\hline
 $\alpha$ & \textbf{AP} & \textbf{AL} & \textbf{DAL} & \textbf{LAAL} & \textbf{BLEU} \\ \hline

0.8                  & 0.80        & 705        & 1973        & 1289         & 14.43    \\
0.7                  & 0.82        & 1287       & 2430        & 1765         & 15.93    \\
0.6                  & 0.86        & 1996       & 3009        & 2362         & 17.57     \\
0.5                  & 0.89        & 2897       & 3736       &  3152        &  19.87     \\
0.4                  & 0.93        & 4045       & 4562       &  4149       &   22.53  \\   
0.3                  & 0.97        & 4947       & 5198       &  4971        &  23.97   \\
0.2                  &  0.99    & 5460         &   5540      & 5463         & 24.54    \\
0.1                  & 0.99     & 5636         & 5643        & 5636         & 24.77 \\
0                 & -        & -        & -        & -         & 25.39         \\

\hline
\end{tabular}
    \caption{Numerical results of EDAtt on MuST-C English to German speech-to-text translation dataset.}
\label{table:ende_edatt}
\end{table*}

\begin{table*}[h]
\centering
\begin{tabular}{c|cccc|c} \hline
\multicolumn{6}{c}{\textit{\textbf{EDAtt on En$\rightarrow$Es}}}                                  \\\hline
 $\alpha$ & \textbf{AP} & \textbf{AL} & \textbf{DAL} & \textbf{LAAL} & \textbf{BLEU} \\ \hline

0.8                 &  0.81	 & 715   	&   1939    &	1184    &    22.93      \\
0.7                 &  0.82  & 900      &	2119    &	1319    &    24.19    \\
0.6                 &  0.84	 & 1104	    &   2314    &	1491    &    25.15      \\
0.5                 &  0.85	 & 1321	    &   2489	&   1661    &    26.31      \\
0.4                 &  0.87	 & 1547     &	2688    &	1855    &    27.02   \\
0.3                 &  0.89	 & 1822      &	2939    &	2089    &    27.81      \\
0.2                 &  0.92  &	2328    &	3454    &	2554    &    28.42     \\
0.1                 &  1.00	& 3853     &	4770   &	3984    &    29.11       \\
0                 &     -    &   -      &    -     &    -      &   31.20      \\

\hline
\end{tabular}
\caption{Numerical results of EDAtt on MuST-C English to Spanish speech-to-text translation dataset.}
\label{table:enes_edatt}
\end{table*}

\subsection{Speech-to-Speech}
\label{app:numerical_s2s}
\begin{table*}[h]
\centering
\begin{tabular}{c|ccccc|c} \hline
\multicolumn{7}{c}{\textit{\textbf{NAST-S2S on CVSS-C Fr$\rightarrow$En}}}                                  \\ \hline
 $T_s + T_a (ms)$ & \textbf{AL} & \textbf{AL\_EOW} & \textbf{AL\_BOW} &  \textbf{StartOffset} &	\textbf{EndOffset}   &
 \textbf{ASR-BLEU} \\ \hline
320 + 320 & -393 & 1405 & 1085 & 655 & 562 & 19.67 \\
640 + 640 & 1533 & 1802 & 1455 & 1295 & 863 & 19.15 \\
1280 + 1280 & 3330 & 2961 & 2601 & 2566 & 1648 & 20.20 \\
1920 + 1280 & 3975 & 3390 & 3046 & 3179 & 1920 & 21.77 \\
1920 + 1920 & 4335 & 4021 & 3689 & 3753 & 2292 & 22.70 \\
2560 + 1280 & 4408 & 3785 & 3448 & 3753	& 2175 & 23.58 \\
2560 + 2560 & 4976 & 4886 & 4573 & 4697	& 2753 & 24.88 \\
$\infty$ & - & - & - & - & - & 25.82 \\
 \hline

\end{tabular}
\caption{Numerical results of NAST-S2$S$ on CVSS-C French to English speech-to-speech translation dataset.}
\label{table:fren}
\end{table*}

\begin{table*}[h]
\centering
\begin{tabular}{c|cccccc} \hline
\multicolumn{7}{c}{\textit{\textbf{NAST-S2S on CVSS-C Fr$\rightarrow$En}}}                                  \\ \hline
 $T_s + T_a (ms)$ & \textbf{AL} &  \textbf{AL\_CA} &	\textbf{StartOffset}   &
 \textbf{StartOffset\_CA} &  \textbf{EndOffset} &	\textbf{EndOffset\_CA}    \\ \hline
320 + 320 & -393 & 347 & 655 & 713 & 562 & 1550  \\
640 + 640 & 1533 & 1824 & 1295 & 1351 & 863 & 1344 \\
1280 + 1280 & 3330 & 3501 & 2566 & 2642 & 1648 & 1901 \\
1920 + 1280 & 3975 & 4103 & 3179 & 3245 & 1920 & 2088 \\
1920 + 1920 & 4335 & 4482 & 3753 & 3844 & 2291 & 2465 \\
2560 + 1280 & 4408 & 4527 & 3753 & 3823 & 2175 & 2312 \\
2560 + 2560 & 4976 & 5098 & 4697 & 4781 & 2753 & 2879 \\
 \hline

\end{tabular}
\caption{Comparison of non-computation-aware and computation-aware metrics results for NAST-S2$S$ on CVSS-C French to English speech-to-speech translation dataset.}
\label{table:fren_ca}
\end{table*}

\begin{table*}[h]
\centering
\begin{tabular}{c|ccc} 
\hline
\multicolumn{4}{c}{\textit{\textbf{NAST-S2S on CVSS-C Fr$\rightarrow$En}}}                                  \\ \hline

 $ T_s + T_a (ms) $ & \textbf{ASR-BLEU} &  \textbf{ASR-BLEU (Silence Removed)} &	\textbf{AL}  \\ \hline
		
320+320	 & 19.67 &	24.90 &	-393 \\
640+640	& 19.15 &	25.67 &	1533 \\
1280+1280 &	20.20 &	25.71 &	3330 \\
2560+2560 &	24.88 &	26.14 &	4976 \\
\hline 

\end{tabular}
\caption{Comparison between ASR-BLEU and ASR-BLEU (Silence Removed) of NAST-S2$S$ on CVSS-C French to English speech-to-speech translation dataset.}
\label{table:fren_asr_bleu_rm}
\end{table*}

\begin{table*}[h]
\centering
\begin{tabular}{c|ccc} 
\hline
\multicolumn{4}{c}{\textit{\textbf{NAST-S2S on CVSS-C Fr$\rightarrow$En}}}                                  \\ \hline

 $ T_s + T_a (ms) $ & \textbf{ASR-BLEU} &  \textbf{BLASER 2.0} &	\textbf{AL}  \\ \hline
		
320+320	 & 19.67 &	3.022 &	-393 \\
640+640	& 19.15 &	3.017 &	1533 \\
1280+1280 &	20.20 &	3.066 &	3330 \\
1920+1280 &	21.77 &	3.103 &	3975 \\
1920+1920 &	22.70 &	3.113 &	4335 \\
2560+1280 &	23.58 &	3.123 &	4408 \\
2560+2560 &	24.88 &	3.136 &	4976 \\
$\infty$ &	25.82 &	3.144 &	- \\
\hline 
\multicolumn{4}{c}{\textit{\textbf{Offline Models}}} \\ \hline 
S2UT &	23.39	& 3.062 & - \\
UnitY&	27.80 &	3.178 & - \\ \hline

\end{tabular}
\caption{BLASER 2.0 scores of NAST-S2$S$ on CVSS-C French to English speech-to-speech translation dataset.}
\label{table:fren_blaser2}
\end{table*}

In addition to \texttt{AL} and \texttt{AL\_EOW}, we also present results for \texttt{AL\_BOW}, \texttt{StartOffset}, and \texttt{EndOffset}, as measured by the SimulEval toolkit. \texttt{AL\_BOW} is analogous to \texttt{AL\_EOW} but considers the generation time of each word as the beginning time of the corresponding speech. \texttt{StartOffset} and \texttt{EndOffset} measure the offset of the beginning and ending of the generated speech compared with the input speech. We also employ \texttt{BLASER 2.0} to assess the quality of translated speech. The trade-off between latency and quality is attained by adjusting the chunk size $T_s$ and the additional frames $T_a$. The offline results are obtained by setting the chunk size to be longer than any utterance in the dataset ($T_s = \infty$). We use SimulEval \texttt{v1.1.4} for evaluation. The numerical results of NAST-S2$S$ are presented in Table \ref{table:fren}, \ref{table:fren_ca}, \ref{table:fren_asr_bleu_rm} and \ref{table:fren_blaser2}.

\begin{table*}[h]
\centering
\begin{tabular}{ccc|ccccc|c} \hline
\multicolumn{9}{c}{\textit{\textbf{Wait-k-Stride-n on CVSS-C Fr$\rightarrow$En}}}                                  \\ \hline
$T_s (ms)$ & $n$ & $5$ & \textbf{AL} &  \textbf{StartOffset} &	\textbf{EndOffset}   & \textbf{DCNum} & \textbf{DCAve} & \textbf{ASR-BLEU} \\ \hline
320 & 5 & 5  &  -164  &  1934  &   1503    &     11.7   &  161 &  \enspace 8.41 \\
320 & 5 & 10 &  2154  &  3472  &   2172    &     6.9    &  136 &  13.30 \\
320 & 5 & 15 & 4023   &  4697  &   2766	    &     3.1   &  83 &  17.06 \\
640 & 10 & 1  &  1188 &  1295  &   1242	      &     6.9  &  318 & \enspace 7.34 \\
640 & 10 & 3  &  2449  &  2566  &   1731    &    4.9     &  294 &  11.61 \\
640 & 10 & 5  &  3627  &  3753  &   2312    &     3.0    &  235 &  14.55 \\
1280 & 20 & 1  & 3302   &  2566   &   1693    &    2.5   &  541 &  14.06 \\
1280 & 20 & 2  &  4159  &  3753  &   2248    &     1.5   &  404 &  16.18 \\
1280 & 20 & 3  & 4859  &  4697  &    2732    &     0.8   &  233 &  17.91 \\
\hline

\end{tabular}
\caption{Numerical results of Wait-$k$-Stride-$n$ on CVSS-C French to English speech-to-speech translation dataset.}
\label{table:fren_waitk}
\end{table*}

\begin{table*}[h]
\centering
\begin{tabular}{c|ccccc|c} \hline
\multicolumn{7}{c}{\textit{\textbf{EDAtt + Tacotron2 on CVSS-C Fr$\rightarrow$En}}}                                  \\ \hline
$\alpha$ & \textbf{AL} &  \textbf{StartOffset} &	\textbf{EndOffset}   & \textbf{DCNum} & \textbf{DCAve} & \textbf{ASR-BLEU} \\ \hline
0.8  & 2850 & 2131 & 5846 & 0.8 & 360 & 11.90\\
0.6  &  3136  &  2383  & 5451    &    0.8     &   442 &  13.69 \\
0.4  & 3585   &  2859  &   4848	    &     0.7   &  472 &  15.93 \\
0.2  & 4431  & 3922 &	3887     &    0.4 & 358 & 19.76  \\

\hline

\end{tabular}
\caption{Numerical results of EDAtt + Tacotron2 on CVSS-C French to English speech-to-speech translation dataset.}
\label{table:fren_edatt}
\end{table*}

\section{Analysis on Length Ratio}
We present the ablation study of model hyperparameter $r_{\mathrm{down}}$ and $r_{\mathrm{up}}$ in Table \ref{table:ab_downsample} and \ref{table:ab_upsample}.
\label{app:analysis_ratio}
\begin{table}[t]
\centering
\begin{tabular}{c|ccc}
\toprule
  $r_{\mathrm{down}}$           &  $\mathbf{1}$  & $\mathbf{2}$ & $\mathbf{4}$ \\ 
\midrule
  $L_{decoder} / L_{target} $  & 9.3  & 4.6 &  2.3  \\
  \textbf{BLEU}  & 24.52  & 24.54 & 22.05  \\
\bottomrule
\end{tabular}
\caption{
Performance of offline NAST-S2$T$ with varying hyperparameter $r_{\mathrm{down}}$ on MuST-C English to German speech-to-text translation dataset. $L_{decoder}$ and $L_{target}$ represent the length of linguistic decoder and text target, respectively. The average ratio of these lengths is calculated using the training dataset.}
\label{table:ab_downsample}
\end{table}

\begin{table}[t]
\centering
\begin{tabular}{c|ccc}
\toprule
  $r_{\mathrm{up}}$           &  $\mathbf{4}$  & $\mathbf{6}$ & $\mathbf{8}$ \\ 
\midrule
   $L_{decoder} / L_{target} $  & 2.4  & 3.6 &  4.8  \\
  \textbf{ASR-BLEU}  & 25.06  & 25.82 & 26.16  \\
\bottomrule
\end{tabular}
\caption{Performance of offline NAST-S2$S$ with varying hyperparameter $r_{\mathrm{up}}$ when $r_{\mathrm{down}}$ is fixed to $2$ on CVSS-C French to English speech-to-speech translation dataset. $L_{decoder}$ and $L_{target}$ represent the length of acoustic decoder and unit target, respectively. The average ratio of these lengths is calculated using the training dataset.}
\label{table:ab_upsample}
\end{table}

\end{document}